\documentclass[letterpaper]{article} 
\usepackage{aaai2026}  
\usepackage{times}  
\usepackage{helvet}  
\usepackage{courier}  
\usepackage[hyphens]{url}  
\usepackage{graphicx} 
\usepackage{natbib}  
\usepackage{caption} 
\usepackage{algorithm}
\usepackage{algorithmic}

\usepackage{newfloat}
\usepackage{listings}
\DeclareCaptionStyle{ruled}{labelfont=normalfont,labelsep=colon,strut=off} 
\floatstyle{ruled}
\newfloat{listing}{tb}{lst}{}
\floatname{listing}{Listing}
\usepackage{soul}
\usepackage[most]{tcolorbox}
\usepackage{chemformula}
\usepackage{pgfplotstable}
\usepackage{colortbl}
\usepackage{amssymb}
\usepackage{siunitx}
\usepackage{tikz}
\usetikzlibrary{shapes.geometric}
\usepackage{subcaption}
\usepackage{booktabs}
\usepackage{url}
\usepackage{multirow}

\newtheorem{definition}{Definition}

\let\origthelstnumber\thelstnumber
\makeatletter
\newcommand*\Suppressnumber{%
  \lst@AddToHook{OnNewLine}{%
    \let\thelstnumber\relax%
    \advance\c@lstnumber-\@ne\relax%
  }%
}
\newcommand*\Reactivatenumber[1]{%
  \setcounter{lstnumber}{\numexpr#1-1\relax}
  \lst@AddToHook{OnNewLine}{%
    \let\thelstnumber\origthelstnumber%
    \refstepcounter{lstnumber}
  }%
}
\makeatother

\definecolor{myorange}{RGB}{211,140,90}
\newcommand{\hexnum}[1]{%
  \tikz[baseline=(X.base)]{
    \node[
      draw=myorange, fill=myorange, text=white,
      regular polygon, regular polygon sides=6,
      minimum size=\ht\strutbox, 
      inner sep=1pt
    ] (X) {\sffamily\bfseries #1};
  }%
}

\newcommand*{\Comment}[1]{\hfill\mbox{\textit{#1}}}
\newtcolorbox{cleanframebox}{
  enhanced,
  colframe=black,
  colback=white,
  boxrule=0.4pt,
  arc=0pt,
  outer arc=0pt,
  boxsep=3pt,
  left=0pt,
  right=0pt,
  top=0pt,
  bottom=0pt,
  overlay={
    \draw[line width=0.4pt,black]
    ([yshift=-3.2pt,xshift=3.2pt]frame.south west) --
    ([yshift=-3.2pt,xshift=0pt]frame.south east);
  }
}

\title{Failure Detection in Chemical Processes Using Symbolic Machine Learning: A Case Study on Ethylene Oxidation}
\author{
  Julien Amblard\textsuperscript{\rm 1},
  Niklas Groll\textsuperscript{\rm 2},
  Matthew Tait\textsuperscript{\rm 3},
  Mark Law\textsuperscript{\rm 3},
  Gürkan Sin\textsuperscript{\rm 2},
  Alessandra Russo\textsuperscript{\rm 1}
}
\affiliations{
  \textsuperscript{\rm 1}Department of Computing, Imperial College London, London, United Kingdom\\
  \textsuperscript{\rm 2}Department of Chemical and Biochemical Engineering, Technical University of Denmark, Kongens Lyngby, Denmark\\
  \textsuperscript{\rm 3}ILASP Limited, London, United Kingdom\\
  julien.amblard16@imperial.ac.uk, nigro@kt.dtu.dk
}

\usepackage{bibentry}

\usepackage{fancyhdr}
\nocopyright

\begin{document}

\maketitle

\begin{abstract}
  Over the past decade, Artificial Intelligence has significantly advanced, mostly driven by large-scale neural approaches. However, in the chemical process industry, where safety is critical, these methods are often unsuitable due to their brittleness, and lack of explainability and interpretability. Furthermore, open-source real-world datasets containing historical failures are scarce in this domain.

  In this paper, we investigate an approach for predicting failures in chemical processes using symbolic machine learning and conduct a feasibility study in the context of an ethylene oxidation process. Our method builds on a state-of-the-art symbolic machine learning system capable of learning predictive models in the form of probabilistic rules from context-dependent noisy examples. This system is a general-purpose symbolic learner, which makes our approach independent of any specific chemical process. To address the lack of real-world failure data, we conduct our feasibility study leveraging data generated from a chemical process simulator. Experimental results show that symbolic machine learning can outperform baseline methods such as random forest and multilayer perceptron, while preserving interpretability through the generation of compact, rule-based predictive models. Finally, we explain how such learned rule-based models could be integrated into agents to assist chemical plant operators in decision-making during potential failures.
\end{abstract}

\begin{links}
  \link{Supplementary Material}{https://github.com/julambl75/ethylene-oxidation-ilp-results}
  \link{Code}{https://github.com/julambl75/ethylene-oxidation-ilp}
\end{links}

\fancypagestyle{firstpage}{
  \fancyhf{}
  \fancyfoot[C]{Preprint. Accepted at AAAI-MAKE 2026.}
  \renewcommand{\headrulewidth}{0pt}
}
\thispagestyle{firstpage}

\section{Introduction}

The chemical process industry is one of the world's largest industrial sectors, producing a wide range of products from materials to consumer goods. Though it is important that industrial processes operate efficiently, it is paramount that they operate safely, as unsafe operating conditions can have catastrophic repercussions\footnote{Notable incidents include the Seveso (1976)~\citep*{eskenazi2018seveso} and Bhopal (1984)~\citep*{broughton2005bhopal} disasters, which both involved the leakage of dangerous chemicals, incurring long-term damage to the local populations and environment. In both cases, insufficient safety measures were to blame, combined with a sequence of operational failures that eventually led to the incident.}.

Chemical processes often involve substances under high pressure and temperature; these can be toxic, flammable, or even explosive when mixed with oxygen. For this reason, chemical process plants must follow a set of strict safety regulations (varying between continents and countries) in all phases of their lifetimes. One crucial part is the hazard and operability (HAZOP) study~\citep*{crawley2015hazop}, which identifies possible failures and their potential consequences — a comprehensive HAZOP study is mandatory before a plant starts operating and must be revised and updated every few years.

To ensure safe operating conditions during the use phase, processes are constantly monitored by a team of human operators, who receive data from online sensors (i.e., real-time data) and offline measurements (i.e., resulting from analyzing a sample in an offsite laboratory). Based on these data, their education and training, and any underlying protocols, the human operators intervene to maintain process safety. Typically, simple tools are used to help the operators review and react to changes in the data, the most common of which are live alarms that launch when the value of a process variable reaches a predefined threshold.

In a world where chemical processes are becoming ever more complex, the automation of failure detection has been an ongoing area of research~\citep*{taqvi2021review}. While early approaches were purely symbolic, such as the logic-based \textit{expert system} MODEX~\citep*{rich1987model}, recent work in this field has focused on classical machine learning (ML) algorithms~\citep*{thon2021artificial, plathottam2023review}. However, symbolic solutions generally require considerable manual engineering and often scale poorly, while ML approaches lack explainability and interpretability, making them difficult to trust. Moreover, real-world process failure data are not readily available, and even when they are data sparsity is common: this is one of the major challenges in chemical process safety. For an analogy, consider airline accidents — flight data for a particular failure mode can only exist if the failure has occurred before, and even then may be incomplete.

In this paper, we propose a methodology based on symbolic ML for failure detection in chemical processes. The key challenge is to learn interpretable models for predicting failures from (past or simulated) failure scenarios. We make use of DisPLAS, a system for approximate probabilistic logic-based Learning from Answer Sets (LAS) that uses a restricted version of the Learning from Noisy Answer Sets~\cite{law2020fastlas} and optimizes the posterior probability of the learned rules by discretizing the space of possible probabilities. We conduct a feasibility study of our approach present based on the ethylene oxidation process. Using a widely-adopted industrial chemical process simulator~\cite{boys2023architecture}, we generate failure data and use these to learn the cause-effect relations at play in the process and the patterns associated with different failures. Rules are learned by means of DisPLAS. As far as we are aware, data-driven logic-based learning techniques have never been applied to a chemical process safety setting for classifying failures.

We evaluate the performance of our framework against ML baselines such as multilayer perceptron and random forest, and show that our approach outperforms while generating more interpretable models. We then discuss the relevance and quality of generated rules, and outline an agentic framework which could be used to provide an AI-in-the-loop solution for assisting chemical plant operators in identifying process failures. Finally, we elaborate on the implications of this case study, and outline future research directions.

\section{Background}

DisPLAS~\cite{law2025discretised} is an approximate probabilistic symbolic machine learning approach for optimizing the posterior probability of learned rules. It is based on a restricted version of the learning system FastLAS~\cite{law2020fastlas} capable of learning answer set programs (ASP) from noisy data. Answer set programs~\cite{lifschitz2019answer} support default negation. Given atoms $h, b_1, \ldots, b_n, c_1, \ldots, c_m$, a \emph{normal rule} is of the form $h \leftarrow b_1, \dots, b_n, \text{not}\, c_1, \dots, \text{not}\, c_m$, where $h$ is the \emph{head}, $b_1, \ldots, b_n, \text{not}\, c_1, \ldots, \text{not}\, c_m$ (collectively) make up the \emph{body} of the rule, and ``$\text{not}$'' represents negation-as-failure. In this paper, we assume answer set programs to be sets of normal rules. The Herbrand base of a program $P$, denoted $HB_P$, is the set of all (variable-free) atoms formed from predicates and constants in $P$. Given a program $P$ and an interpretation $I\subseteq HB_P$,  the \emph{reduct} $P^I$ is constructed from the grounding (i.e., variable instantiation) of $P$ by removing rules whose bodies contain the negation of an atom in $I$ and removing negative atoms from the remaining rules. An interpretation $I$ is an \emph{answer set} of $P$ iff it is the minimal model of $P^I$.

A symbolic learning system — e.g. ~\cite{muggleton1995inverse} — typically uses an inductive bias, called \emph{mode declarations}, as a form of language bias to specify the search space of possible solutions (called \emph{hypothesis space}) for a given learning task. A \emph{mode bias} is a pair of sets of mode declarations $M =\langle M_h,M_b\rangle$, where $M_h$ (resp. $M_b$) are called the head (resp. body) mode declarations. Given a mode bias $M$, the hypothesis space $S_M$ is the set of normal rules $R$ such that their head and body comply with $M_h$ and $M_b$, respectively.

In this paper we considered a restricted version of the symbolic learner FastLAS~\cite{law2020fastlas}, capable of learning ASP programs from noisy examples. Examples are \emph{weighted context dependent partial
interpretations} (WCDPIs). A partial interpretation $e_{pi}$ is a pair of ground atoms $\langle e^{inc},e^{exc}\rangle$. An interpretation
$I$ \emph{extends} $e_{pi}$ iff $e^{inc}\subseteq I$ and $e^{exc}\cap I =\emptyset$. A WCDPI is
a tuple $e = \langle e_{id},e_{pen},e_{pi},e_{ctx}\rangle$, where $e_{id}$ is an identifier
for $e$, $e_{pen}$ is a positive integer, called a \emph{penalty},
$e_{pi}$ is a partial interpretation and $e_{ctx}$ is an ASP program
called the \emph{context}. A WCDPI $e$ is \emph{accepted} by a program $P$ iff
there is an answer set of $P\cup e_{ctx}$ that extends $e_{pi}$.

A FastLAS learning task is a tuple $T =\langle B,M,E,S\rangle$ where $B$ is an ASP program, $M$ is a \emph{mode bias} that says which predicates can
appear in the head or body of a rule, $E$ is a finite set of WCDPIs and
$S : S_M \rightarrow \mathbb{N}_0$ is a \emph{scoring function}. $B$ defines \textit{background knowledge} that applies task-wide, as opposed to the $e_{ctx}$ which contain knowledge specific to each WCDPI. $S$ assigns a cost (often given by the ``length'' — number of literals) to every rule in the hypothesis space $S_M$. For any hypothesis $H\subseteq S_M$:
\begin{itemize}
  \item For each $e \in E$, $H$ covers $e$ iff $B\cup H$ accepts $e$.
  \item The score of $H$ is defined as $S(H,T)=$
    $\sum_{h\in H} cost(h)$ plus the penalty of
    each example in $T$ not covered by $H$.
  \item $H$ is an optimal solution of $T$ iff
    there is no $H^{'}\subseteq S_M$ such that $S(H^{'},T)<S(H,T)$.
\end{itemize}

DisPLAS~\cite{law2025discretised} is a discretized probabilistic extension of FastLAS, capable of predicting the probability of an event \texttt{ev} in some context $C$; for example, $C$ may be (the range of) a certain variable's values in a chemical plant, and \texttt{ev} could be the event corresponding to a specific failure occurring. For each context $C$, the probability of observing the event \texttt{ev} is modeled by a Bernoulli distribution with parameter $p_C^{\texttt{ev}}$. The goal of DisPLAS is to learn a hypothesis $H$ that predicts $p_{C}^{\texttt{ev}}$ for each context $C$. The mode bias of DisPLAS, denoted as $M_P$, extends that of FastLAS with a set $\Phi$ of probabilities (real numbers in $[0,1]$) and a function $\mathcal{S}$ that defines the prior probability of each rule in the hypothesis space ($\mathcal{S}:S_M \mapsto [0,1]$). Given a FastLAS mode bias $M$, $M_P=\langle M, \Phi,\mathcal{S}\rangle$. The discretized probabilistic hypothesis space $S_{M_{P}}$ is the set of rules of the form $prob_{\phi}\leftarrow b_1,...,b_n$, where $\phi \in \Phi$ and each head and body literal is compatible with $M$ (i.e., with $M_h$ and $M_b$).

\begin{definition}
  A \textit{discretized probabilistic} LAS (DisPLAS) task is a tuple of the form $\langle B, M_P, C^{+}, C^{-}\rangle^\texttt{ev}$, where $B$ is an ASP program called background knowledge, $M_P$ is DisPLAS mode bias, and $C^{+}$
  and $C^{-}$ are sets of ASP programs called (positive and negative, resp.) \textit{contexts}, representing the situations in which an event \texttt{ev} occurs and does not occur (resp.). A solution is a hypothesis $H\subseteq S_M$ that maximizes the posterior probability $P(H\mid\langle C^{+},C^{-}\rangle)$, where
  $\langle C^{+},C^{-}\rangle$ denotes that we have observed event \texttt{ev} in each context in $C^{+}$ and observed the absence of \texttt{ev} in each context in $C^{-}$.
\end{definition}

\noindent
If we now assume the existence of multiple events, each event \texttt{ev} observable in some context $C$ with a probability $p_C^{\texttt{ev}}$ modeled by a Bernoulli distribution, we can see that these events will be independent. Extending the above definition to the multi-event case can therefore be accomplished by decomposing the problem into sub-tasks, each sub-task aimed at finding an optimal hypothesis for predicting a single event (i.e., containing only rules whose head is this event), which maximizes the posterior probability $P(H|\langle C^+,C^-\rangle)$. Intuitively, for a given event \texttt{ev}, we would like a hypothesis that is as much as possible in line with classifying all the $C^{+}$ as \texttt{ev} and not classifying any of the $C^{-}$ as \texttt{ev}. Note that the prior probability of a hypothesis $H\subseteq S_{M_{P}}$ is $\Pi_{h\in H} \mathcal{S}(h) \times
\Pi_{h\in (S_{M_{P}}-H)} (1-\mathcal{S}(h))$.

It can be shown that, for an event \texttt{ev}, $P(H|\langle C^+,C^-\rangle) \propto \prod_{C \in C^+} P(\texttt{ev}|H,C)  \times \prod_{C \in C^-}(1 - P(\texttt{ev}|H,C) ) \times \prod_{h \in H}\frac{\mathcal{S}(h)}{1 - \mathcal{S}(h)}$, where $P(\texttt{ev}|H,C) $ is the \textit{predicted probability}, defined as the largest $\phi \in \Phi$ such that $B \cup C \cup H \models \texttt{ev}_\phi$, under the assumption that the $h \in H$ can be modeled by independent Bernoulli distributions.

In the multi-event case, a DisPLAS task becomes a tuple $\langle B, M_P, E^{+}, E^{-}\rangle$, where $E^+$ (resp. $E^-$) are WCDPIs (with the $e_{inc}$ as singleton sets) instead of contexts $C^+$ (resp. $C^-$) for a single event. The algorithm is applied independently for each event, and the resulting sets of rules are combined into a single hypothesis. Also, the penalties of examples are used as weights when computing the posterior probabilities of hypotheses.

DisPLAS supports integral numeric data types in the contexts, together with binary comparisons over these data types in hypotheses. These are called \textit{numeric variables}.

\section{Methodology}

We now present our approach for learning interpretable models of chemical process failures from failure scenarios, instantiating it in a specific chemical process setting. We consider two DisPLAS learning problems, one focused on the general trends in a process, and the other specialized in failure classification.

\subsection{Chemical Setting and Data Generation}

Ethylene oxidation (EO) is an \textit{exothermic} reaction (i.e., a reaction that releases heat) in which the compound ethylene (\ch{C2H4}) is oxidized to form ethylene oxide (\ch{C2H4O}) in the presence of oxygen: \ch{2 C2H4 + O2 -> 2 C2H4O}~\citep*{lohr2024ethylene}. In mixtures containing oxygen at high enough concentrations, both ethylene and ethylene oxide become flammable, so the product composition\footnote{When referring to the composition of a mixture of gases, we are interested in the respective proportions of its constituents.} should be kept below the flammability limit. The exothermic property of the reaction further increases the risk of combustion, therefore care must be taken to keep the process at a controlled temperature. Failure to do so can lead to a \textit{thermal runaway}, whereby a lack of cooling causes the reaction to speed up and therefore to release more heat, which in turn leads to an even greater reaction rate; eventually, the rate at which the temperature is increasing reaches a point where the process can no longer be kept under control. This can cause anything from a pressure buildup in the system to a rupture or explosion of pipes or of the reactor vessel, and therefore may lead to the release of toxic and/or flammable chemicals — which should be avoided at all cost.

Unfortunately, although there exist reports regarding past incidents involving ethylene oxide (e.g., the release of over 14 tonnes of ethylene oxide in the United States in 2023~\citep{csb2025investigation}), no such report has made available detailed time-series process data. Therefore, we have modeled the EO process using the widely-adopted industrial process simulator AVEVA Process Simulation (APS), which comes built-in with conservation and equilibrium equations based on the laws of thermodynamics. This software has been widely adopted within the chemical process industry, due to its ability to accurately model real-world processes.

A process can be broken down into \textit{components}, which each have a number of \textit{process variables} to describe their state (e.g., for a valve, the opening position). Some process variables can be controlled, in which case we call them \textit{process parameters}. In APS, process parameters are set by the user, and the remaining process variables are computed by the software.

To specify the behaviors of the above-mentioned main reaction (plus two additional minor reactions that also occur), we have integrated the kinetics\footnote{\textit{The reaction kinetics} define a set of equations describing the rates at which one or more reactions occur in a reacting system, taking into account the process conditions (mainly pressure, temperature and fluid composition).} established by \citet*{stoukides1986ethylene}. Together with the software's advanced thermodynamic nonlinear model — whose equations are used to calculate thermochemical properties (e.g., density) — the process conditions can be computed at each point in the simulated system. Processes in APS can be run in multiple modes, two of which are used in our approach:

\begin{itemize}
  \item \textbf{Dynamic} mode operates using differential equations for the conservation of mass and energy. Time is a continuously evolving variable, and any changes made to the operating conditions have gradual effects on the system. Eventually, the simulation either reaches a \textit{steady state} where process variables have stopped evolving with time, or becomes \textit{unsolved} if the underlying equations fail to converge (due to some variables of these equations exceeding their calculation limits).
  \item \textbf{Static} mode is time-invariant and any changes to the operating conditions cause the simulation to immediately move to a new steady (or unsolved) state.
\end{itemize}

\begin{figure}[H]
  \centering
  \includegraphics[width=\linewidth]{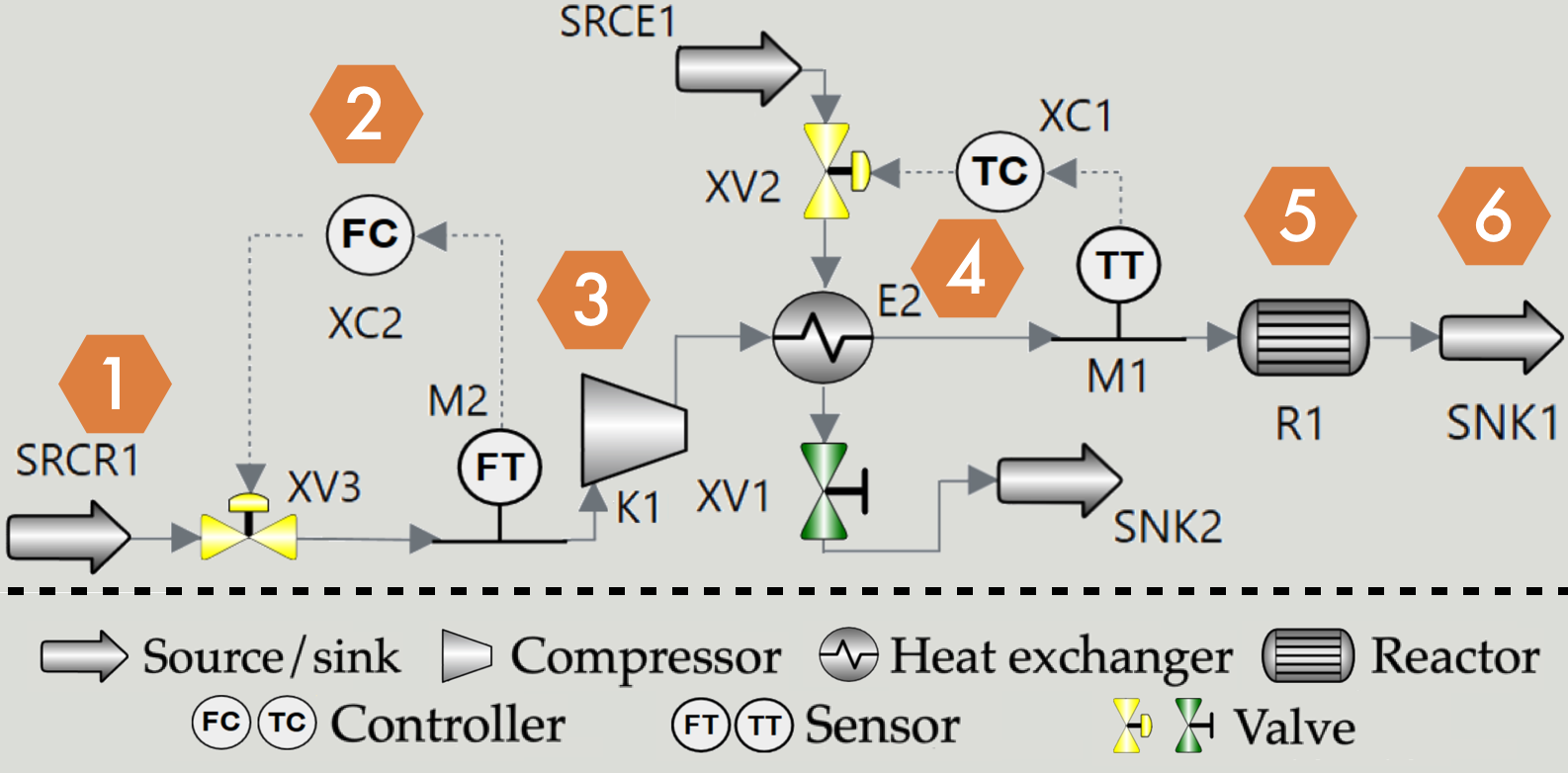}
  \caption{Simulation flowsheet of the EO process, with a legend to identify components}
  \label{fig:process_sim}
\end{figure}

\noindent
Figure~\ref{fig:process_sim} shows the flowsheet for our simulated process, the main steps being as follows: (1) the \textbf{source} serves as the entry point for all gases, (2) the \textbf{flow control loop} regulates the flow rate by controlling the position of its valve, (3) the \textbf{compressor} increases the pressure, (4) the \textbf{heat exchanger} (HX) cools the flow down to a desired reaction temperature by transferring excess heat to a stream of cooling water, (5) the \textbf{reactor} is where the reaction occurs, and (6) the \textbf{sink} is the ``exit node''.

\begin{table}[H]
  \centering
  \begin{tabular}{ccccc}
    \toprule
    \multicolumn{2}{c}{\textbf{Component}} & \multicolumn{2}{c}{\textbf{Process variable}} \\
    \cmidrule(r){1-2} \cmidrule(l){3-4}
    \textbf{Step}
    & \textbf{Name}
    & \textbf{Name}
    & \textbf{Description} \\
    \midrule
    \multirow{2}{*}{\hexnum{1}} & SRCR1 & P & Pressure \\
    & SRCR1 & T & Temperature \\
    \cmidrule{1-4}
    \hexnum{2} & M2 & PV & Measured flow rate \\
    \cmidrule{1-4}
    \multirow{2}{*}{\hexnum{3}} & K1 & P1 & Inlet pressure \\
    & K1 & P2 & Outlet pressure \\
    \cmidrule{1-4}
    \multirow{3}{*}{\hexnum{4}} & M1 & PV & Measured temperature \\
    & E2 & Tsi & Cooling water inlet temp. \\
    & E2 & Tti & Main stream inlet temp. \\
    \cmidrule{1-4}
    \hexnum{5} & R1 & T2 & Outlet temperature \\
    \cmidrule{1-4}
    \multirow{2}{*}{\hexnum{6}} & SNK1 & P & Pressure \\
    & SNK1 & T & Temperature \\
    \bottomrule
  \end{tabular}
  \captionof{table}{The main process variables of the EO process we focus on in this paper — entries in the first column correspond to the steps from Figure~\ref{fig:process_sim}, and the units of the process variables are respectively bar, ºC and kg/s for pressure, temperature and flow rate}
  \label{tab:outcome_params}
\end{table}

Typically, the online sensors found in industrial processes are limited to temperature, pressure, level (the height of a fluid in a vessel), and sometimes flow rate. In this paper, we fill focus primarily on a subset of the process variables made available in the simulation; these are listed in Table~\ref{tab:outcome_params}.

We have defined a set of \textit{experiments}, each one focused on introducing a specific failure into the system by perturbing one or more process variables, assigning new values by sampling from uniform distributions. A small amount of noise is added to unperturbed process variables in dynamic mode, to more closely resemble real-world conditions. This can also act as a proxy for potential oscillations. For every mode-experiment pair (in addition to the ``nominal'' case, where no failure is introduced), we have created a ``perturbation file'' which specifies, for various process variables, a valid range from which to sample uniformly when selecting the value to set, as shown in Figure~\ref{fig:sim_files}. The format also allows toggling whether a process variable should be specified (e.g., to simulate a valve that is stuck closed).

\pgfplotstableread[col sep=comma]{
  param,unit,default,min,max,instr
  XC1.SP,bool,uncheck,,,
  XC1.OP,bool,check,,,
  XC1.OP,fraction,0,0,0,
}\simfilestuckopentempcontrolvalve

\pgfplotstableread[col sep=comma]{
  param,unit,default,min,max,instr
  SRCR1.P,bar,2,1.6,1.98,uniform
  SRCE1.P,bar,2,1.95,2.05,uniform
  SRCR1.M[C2H4],kmol,4464,4240,4360,uniform
}\simfilestaticlowpressuresource

\begin{figure}[H]
  \centering
  \centering
  \setlength{\tabcolsep}{4pt} 
  \pgfplotstabletypeset[
    columns/param/.style={
      string type,
      column type/.add={>{\columncolor{red!10}}}{},
      postproc cell content/.style={
        @cell content/.add={\cellcolor{red!20}}{}
      }
    },
    columns/unit/.style={
      string type,
      column type/.add={>{\columncolor{orange!10}}}{},
      postproc cell content/.style={
        @cell content/.add={\cellcolor{orange!20}}{}
      }
    },
    columns/default/.style={
      string type,
      column type/.add={>{\columncolor{yellow!10}}}{},
      postproc cell content/.style={
        @cell content/.add={\cellcolor{yellow!20}}{}
      }
    },
    columns/min/.style={
      string type,
      column type/.add={>{\columncolor{green!10}}}{},
      postproc cell content/.style={
        @cell content/.add={\cellcolor{green!20}}{}
      }
    },
    columns/max/.style={
      string type,
      column type/.add={>{\columncolor{cyan!10}}}{},
      postproc cell content/.style={
        @cell content/.add={\cellcolor{cyan!20}}{}
      }
    },
    columns/instr/.style={
      string type,
      column type/.add={>{\columncolor{blue!10}}}{},
      postproc cell content/.style={
        @cell content/.add={\cellcolor{blue!20}}{}
      }
    },
    every head row/.style={
      before row=\toprule, after row=\midrule
    },
    every last row/.style={
      after row=\bottomrule
    }
  ]{\simfilestaticlowpressuresource}
  \caption{A perturbation file for ``low pressure at source''}
  \label{fig:sim_files}
\end{figure}

Using these perturbation files, data from 125 \textit{runs} were collected for each experiment using the API provided by APS, both in static and dynamic mode. In dynamic mode, we first recorded the initial state, then applied perturbations, and only then collected data at the following timepoints (in s): 2, 4, 6, 8, 10, 20, 30, 40, 60, 80, 100\footnote{Naturally, the more time goes by after a perturbation is made, the more the process will deviate from normal operating conditions. Collecting data from various timepoints is a step towards finding the right compromise between detecting abnormal behavior early and correctly classifying abnormal behavior.}. The dimensions of the resulting datasets are 27x1501 and 28x2700 for static and dynamic mode respectively\footnote{The columns are: the failure being simulated, the index of the run, the timepoint (only for dynamic mode), and the 25 process variables we chose to monitor. Each row beneath the header is effectively a tuple of values describing process conditions.}.

\subsection{Knowledge Representation and Task Specification}

For the considered process variables $\mathcal{V}$, we define $2k+1$ ($k \in \mathbb{N}_0$) \textit{buckets}. These provide a discrete partition over the range of possible values for each process variable. Buckets are defined through boundaries that mark the transition from one bucket to the next, and these boundaries are closer together for larger $k$. A separate set of boundaries is defined for each ``type'' of process variable (e.g., pressure), as some may have a much greater influence than others on the system when a deviation occurs. If we assume a world with three buckets then we could use, for a given process variable with nominal value $x$, the range $(-\infty, 0.5x)$ for bucket \texttt{low}, $[0.5x, 1.5x]$ for \texttt{normal}, and $(1.5x, \infty)$ for \texttt{high}.

Given the two modes in which APS can run processes, we define DisPLAS learning tasks $T_{static}$ and $T_{dynamic}$.

$T_{static}$ predicts downstream\footnote{Given a reference point in a flowsheet, \textit{downstream} refers to locations that fluids at the reference point can flow to, according to the direction of flow (antonym: \textit{upstream}).} variations from normal behavior given an observed deviation or failure. Specifically, it learns rules about the long-term increase or decrease of process variable values wrt. abnormal process conditions.

In the next definition, $\mathcal{B}$ and $\mathcal{F}$ refer to the set of buckets and failures considered (resp.). For a given process variable $p$ belonging to component $c$, $upstr(v)$ is the set of process variables in $\mathcal{V}$ from components that are upstream of $c$. $value_{v}$ denotes the value of $v$ in the context being looked at, and $v(value_{v})$ is syntax for associating process variable $v$ with its value. Finally, $num\_var(v)$ indicates a numeric variable with a range given by the extreme values of $v$ over the entire dataset.

\begin{definition}
  Recall that a DisPLAS task can be written as a tuple of the form $\langle B, M_P, E^{+}, E^{-}\rangle$. $T_{static}$ is the DisPLAS task $\langle \emptyset, \langle \langle M_h, M_b \rangle, \Phi, \mathcal{S} \rangle, E^{+}, \emptyset \rangle$, where $\mathcal{S}$ is the scoring function implemented by the solver, and:
  \[
    M_h = \{ v(b) \mid \forall v \in \mathcal{V}, b \in \mathcal{B} \},
  \]
  \[
    M_b = \mathcal{F} \cup \{ num\_var(v) \mid \forall v \in \mathcal{V} \},
  \]
  \[
    \Phi = [0.1,0.2,...,1],
  \]
  and each $e \in E^+$ focuses on one run of the experiment concerned with failure $f$, with a chosen ``reference'' process variable $v$ whose value fits into bucket $b_{p}$, such that,
  \[
    e_{id} = index(e), \qquad
    e_{pen} = 100,
  \]
  \[
    e^{inc} = \{ v(b_v) \}, \qquad
    e^{exc} = \{ v(b) \mid \forall b \in \mathcal{B}, b \neq b_{v} \},
  \]
  \[
    e_{ctx} = \{ f \} \cup \{ v'(value_{v'}) \mid \forall v' \in upstr(v) \},
  \]
\end{definition}

\noindent
$T_{dynamic}$ aims at learning failure classification rules based on observations made at $t = t_{short\_term}$, shortly after a failure occurs. It is a multi-label multi-class classifier that predicts the most likely failures given a set of process conditions, as different failures can lead to similar system states.

In the next definition, $v^\Delta$ refers to the change in the discretized value of a process variable $v$ between $t = 0$ and $t = t_{short\_term}$: $v^{\Delta=\nearrow}$ denotes a positive change and $v^{\Delta=\searrow}$ a negative change; $v^{\Delta=\approx}$ is used otherwise. $change\_atom_v$ is used as a placeholder for the applicable change in value in a context $v^{\Delta=\nearrow}(value_{v^{\Delta=\nearrow}})$ or $v^{\Delta=\searrow}(value_{v^{\Delta=\searrow}})$, or lack thereof $v^{\Delta=\approx}$.

\begin{definition}
  $T_{dynamic}$ is the learning task defined as $\langle \emptyset, \langle \langle M_h, M_b \rangle, \Phi, \mathcal{S} \rangle, E^{+}, \emptyset \rangle$, where $\Phi$ and $\mathcal{S}$ are the same as in $T_{static}$, and:
  \[
    M_h = \mathcal{F},
  \]
  \begin{multline*}
    M_b = \{ num\_var(v), v^{\Delta=\approx} \mid \forall v \in \mathcal{V} \} \;\cup \\ \{ num\_var(v^{\Delta=d}) \mid \forall v \in \mathcal{V}, d \in \{ \nearrow, \searrow \} \},
  \end{multline*}
  and each $e \in E^+$ corresponds to a run in which failure $f$ has occurred, such that,
  \[
    e_{id} = index(e),
  \]
  \[
    e_{\text{pen}} = 225 \;\text{if } f = \text{null},\; 100 \;\text{otherwise},
  \]
  \[
    e^{inc} = \{ f \}, \qquad
    e^{exc} = \{ f' \mid \forall f' \in \mathcal{F}, f' \neq f \},
  \]
  \[
  e^{ctx} = \bigcup_{v \in \mathcal{V}} \{ v(value_v), change\_atom_v) \},
\]
\end{definition}

\noindent
Note that although we always generate the same number of training examples for each class, a higher penalty is assigned for those of the ``nominal class'' (i.e., the experiment where no failure is introduced). This was done because normal operating conditions are most common, and to avoid generating rules that are too sensitive to small deviations.

In both learning tasks, process variables have their measurements multiplied by a power of 10, so as to put all numeric variable arguments on a similar scale. These multipliers are determined based on the minimum and maximum values over the example contexts, for each process variable.

We use bias constraints to restrict the search space (e.g., for $T_{static}$, we prune out rules that are entirely about nominal conditions), and adjust the built-in scoring function to penalize rule heads more than rules bodies. In doing so, we generate fewer, more precise rules.

The Supplementary Material contains examples of LAS tasks provided to DisPLAS, for both $T_{static}$ and $T_{dynamic}$.

We considered various \textit{learning parameters}, including:

\begin{itemize}
\item The \textbf{learning task}, either $T_{static}$ or $T_{dynamic}$.
\item The \textbf{experiments} to run i.e., which failures to reproduce.
\item The \textbf{number of runs} to use from each experiment: small values such as 10 encourage fast solving, while much larger values should lead to higher accuracy (\underline{default:} 75).
\item For $T_{dynamic}$, the \textbf{short-term timepoint} $t_{short\_term}$: as time passes, it becomes more and more obvious what the state of the simulation is evolving towards, and therefore easier to link these trends with a cause. We would like to determine the smallest possible value for $t_{short\_term}$ such that failures can be accurately identified, and therefore $t_{short\_term}$ defines the timepoint of the data we include in the contexts of examples in $T_{dynamic}$ (\underline{default:} 6s).
\item \textbf{Restricting the process variables} that can appear in the rule bodies (\underline{default:} only the ``real-world'' process variables, which are shown in Figure~\ref{tab:outcome_params}).
\item The \textbf{proportion of training examples} to reserve for \textbf{cross-validation} (\underline{default:} 20\%).
\end{itemize}

\section{Results}

\begin{table*}[!t]
\centering
\begin{tabular}{cccccccc}
  \toprule
  & \multicolumn{1}{c}{\textbf{Learning parameters}} & \multicolumn{2}{c}{\textbf{Accuracy}} & \multicolumn{2}{c}{\textbf{Interpretability}} & \multicolumn{1}{c}{\textbf{Confidence}} \\
  \cmidrule(r){2-2} \cmidrule(l){3-4} \cmidrule(l){5-6} \cmidrule(l){7-7}
  & Non-default assignments
  & $min(auc)$
  & $avg(auc)$
  & $avg(n_{body})$
  & $avg(n_{r}|cl)$
  & $avg(prob)$ \\
  \midrule
  $$ (1) $$ & \( exp := trivial \) & \textbf{1} & \textbf{1} & \textbf{2.14} & 1.4 & 0.74 \\
  $$ (2) $$ & \( exp := nontrivial6 \) & 0.82 & 0.92 & 3 & \textbf{1.33} & 0.55 \\
  $$ (3) $$ & \( exp := nontrivial10 \) & 0.75 & 0.88 & 3 & 1.7 & \textbf{0.75} \\
  $$ (4) $$ & \( exp := nontrivial10;~n\_runs := 125 \) & 0.7 & 0.85 & 3.11 & 1.8 & 0.73 \\
  \cmidrule{1-7}
  $$ (5) $$ & \( n\_runs := 10 \) & \textbf{0.83} & 0.93 & \textbf{2.5} & \textbf{1} & \textbf{0.7} \\
  $$ (6) $$ & \( n\_runs := 25 \) & \textbf{0.83} & \textbf{0.94} & 2.67 & \textbf{1} & \textbf{0.7} \\
  $$ (7) $$ & \( n\_runs := 75 \) & 0.82 & 0.92 & 3 & 1.33 & 0.55 \\
  $$ (8) $$ & \( n\_runs := 125 \) & 0.75 & 0.87 & 2.8 & 1.67 & 0.53 \\
  \cmidrule{1-7}
  $$ (9) $$ & \( short\_term := 2 \) & 0.67 & 0.81 & 3.18 & 1.83 & 0.64 \\
  $$ (10) $$ & \( short\_term := 4 \) & 0.83 & 0.92 & 3.14 & \textbf{1.17} & \textbf{0.74} \\
  $$ (11) $$ & \( short\_term := 6 \) & 0.86 & 0.93 & 3 & 1.83 & 0.6 \\
  $$ (12) $$ & \( short\_term := 10 \) & 0.86 & 0.93 & 3.27 & 1.83 & 0.6 \\
  $$ (13) $$ & \( short\_term := 20 \) & \textbf{0.93} & \textbf{0.98} & \textbf{3} & 3 & 0.66 \\
  \cmidrule{1-7}
  $$ (14) $$ & \( proc\_vars := all \) & \textbf{0.92} & \textbf{0.94} & \textbf{1.67} & \textbf{1} & \textbf{1} \\
  $$ (15) $$ & \( proc\_vars := real\_world \) & 0.82 & 0.92 & 3 & 1.33 & 0.55 \\
  $$ (16) $$ & \( proc\_vars := m1\_m2 \) & 0.81 & 0.9 & 2.71 & 1.83 & 0.62 \\
  \bottomrule
\end{tabular}
\captionof{table}{A look at the impacts of adjusting the learning parameters for failure detection with task $T_{dynamic}$, using emphasized text to highlight the results that are preferred under each metric}
\label{tab:main_results}
\end{table*}

We can divide experiments into two main categories: \textit{trivial} experiments where detecting a failure — including in the real world — is equivalent to the failure itself (e.g., the ``low temperature at source'' failure can be detected by measuring the temperature at the source), and \textit{nontrivial} experiments, where failures can only be detected by measuring key process conditions in the local environment (e.g., the composition of gases at the source is typically not measured, so to detect a lack of ethylene, downstream measurements must be used). By default, we restricted learning tasks to six nontrivial experiments: ``missing ethylene at source'', ``low setpoint in flow control loop'', ``low setpoint in temperature control loop'', ``leak before compressor'', ``cooling water pressure low'' and ``cooling water out valve stuck closed''. We chose the six experiments based on their diversity in terms of failure types, and because they are all intuitive enough from a rule interpretation point of view. In addition, the selected failures influence process conditions locally, but also have consequences plantwide, the interpretation of which is generally only obvious to experienced human operators.

\subsection{Validation}

In this paper, we focus on the validation of our approach for failure identification ($T_{dynamic}$). Cross-validation is performed with the train-validate split ratio from the learning parameters. We used the one-vs-rest approach to compute the true positive and false positive rates at different probability levels, generating a receiver operating characteristic (ROC) curve~\citep*{yang2017receiver} for every experiment. The area under the curve (AUC) then becomes a metric to quantitatively express our classifier's diagnostic abilities.

For the proposed methodology, we looked at varying the learning parameters one-by-one. We found that the solve times for $T_{dynamic}$ are typically around 5-10 minutes, peaking at two hours for the largest number of failures, runs and process variables. In terms of metrics, we chose: the minimum AUC $min(auc)$, the AUC averaged over the selected failure classes $avg(auc)$, the average rule body length $avg(n_{body})$, the average number of rules per failure class $avg(n_{r}|cl)$, and the average rule probability $avg(p)$. Table~\ref{tab:main_results} shows the results of this investigation, where: $exp$ refers to the experiments included in the task, $n\_runs$ is the number of examples per failure class, $short\_term$ is the value of $t_{short\_term}$, and $proc\_vars$ are the process variables included as features in the examples (either all of them, those from Table~\ref{tab:outcome_params} or just M1.PV and M2.PV).

From the results, we can draw the following conclusions:
\begin{itemize}
\item The more complicated and numerous the failures, the lower the detection rate.
\item Including more runs does not appear to have a significant impact on accuracy or interpretability.
\item Changing $t_{short\_term}$ does not seem to affect the quality of the rules, though this remains to be confirmed by looking at the rules in detail.
\item Including fewer process variables lowers rule confidence, and accuracy is reduced as more are excluded. However, the accuracy is still fairly high with only M1 and M2.
\end{itemize}

\subsection{Baselines}

For $T_{dynamic}$, we compared our logic-based approach to a set of baselines using sklearn~\citep*{pedregosa2011scikit}; these include: support vector machine (SVM), multilayer perceptron (MLP)~\citep*{evans2020evaluating}, random forest (RF), histogram-based gradient boosting (HGB) and adaptive boosting~\citep*{zhu2009multi} (AB). The classifiers were given as training data vectors corresponding to the example contexts of $T_{dynamic}$, with the failure names from the inclusions as target labels; we used the same default train-test split and restrictions on the quantity of data. In addition, hyperparameter optimization was performed using Optuna~\citep*{akiba2019optuna}. Table~\ref{tab:baseline_results} shows the average experiment AUCs for different learning parameter configurations.

\begin{table}[H]
\centering
\begin{tabular}{ccccccc}
  \hline
  Config. & SVM & MLP & RF & HGB & AB & Ours \\ \hline
  $$ (1) $$ & 0.95 & 0.95 & 0.97 & 0.97 & 0.97 & \textbf{1} \\
  $$ (2) $$ & 0.85 & 0.83 & 0.87 & 0.89 & 0.87 & \textbf{0.92} \\
  $$ (3) $$ & 0.84 & 0.77 & 0.85 & 0.84 & 0.85 & \textbf{0.88} \\
  $$ (14) $$ & 0.87 & 0.86 & \textbf{0.97} & \textbf{0.97} & \textbf{0.97} & 0.94 \\
  $$ (15) $$ & 0.85 & 0.83 & 0.87 & 0.89 & 0.87 & \textbf{0.92} \\
  $$ (16) $$ & 0.83 & 0.82 & 0.89 & 0.89 & 0.87 & \textbf{0.9} \\ \hline
\end{tabular}
\caption{The AUC averaged over the selected experiments for different learning parameter configurations (as given in Table~\ref{tab:main_results}), comparing different baselines to our approach}
\label{tab:baseline_results}
\end{table}

\begin{figure*}[t]
\centering
\begin{subfigure}[t]{0.3\linewidth}
  \centering
  \includegraphics[width=\linewidth]{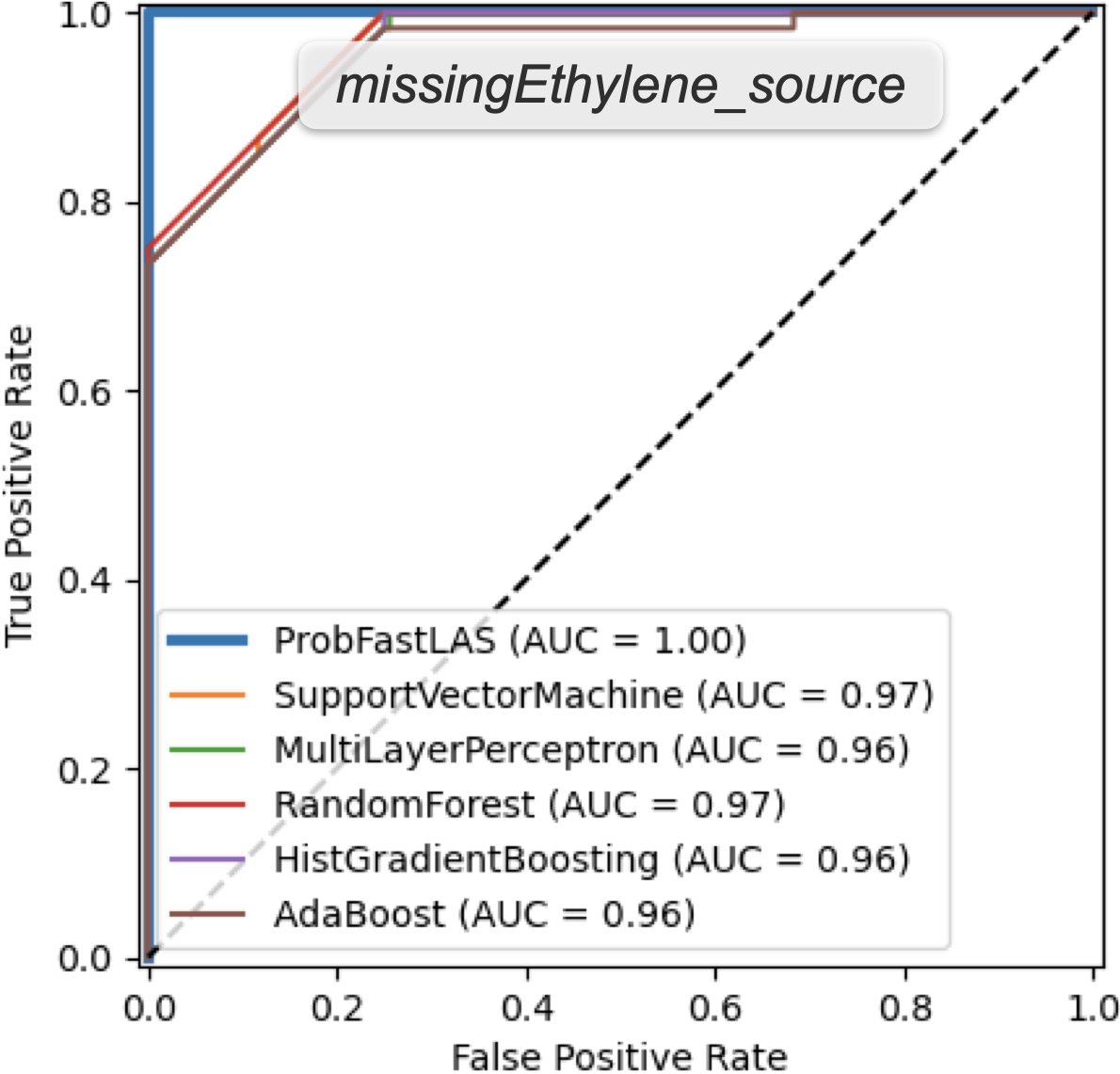}
  \par\vspace{2pt}
  \includegraphics[width=\linewidth]{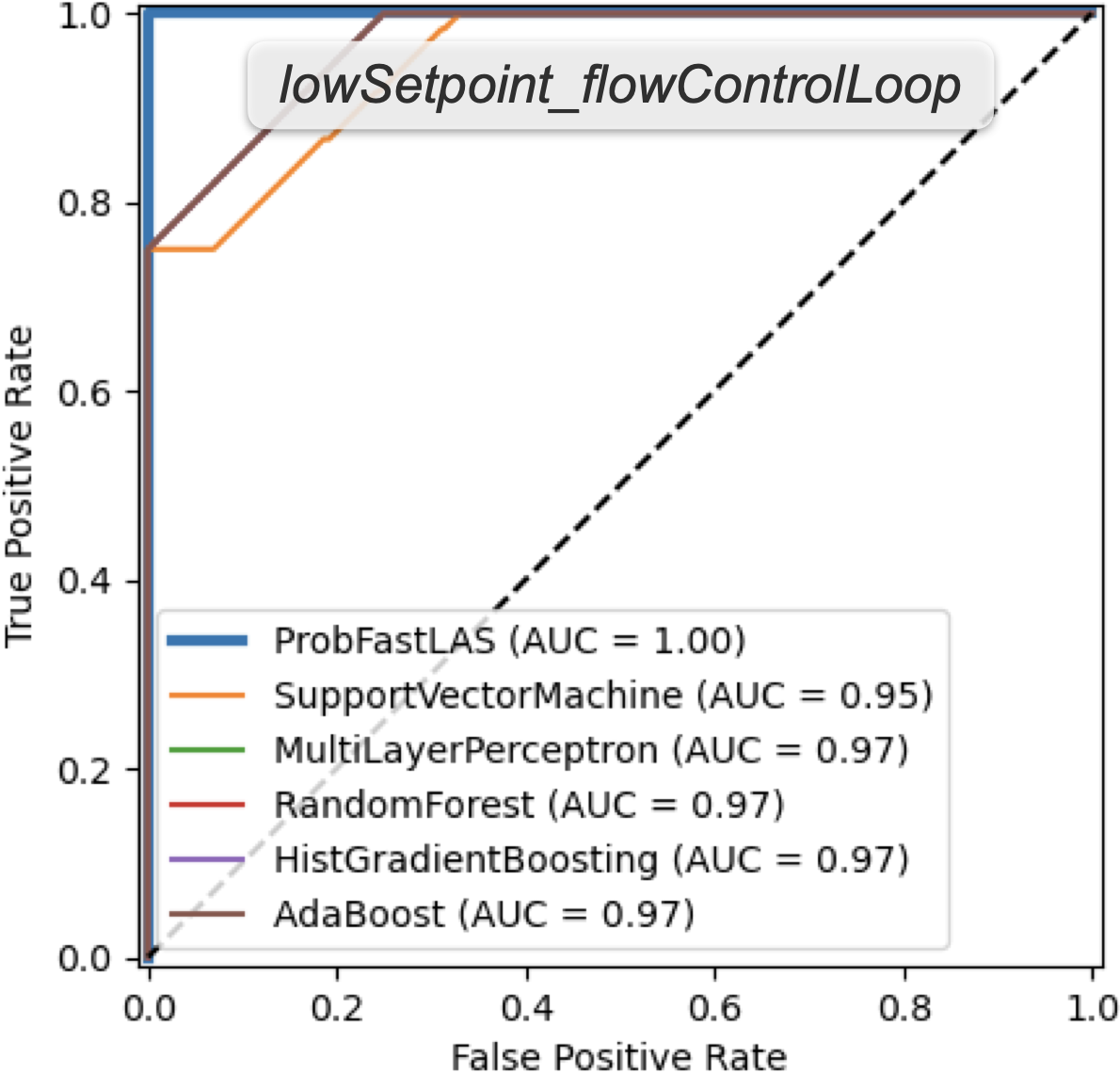}
\end{subfigure}
\begin{subfigure}[t]{0.3\linewidth}
  \centering
  \includegraphics[width=\linewidth]{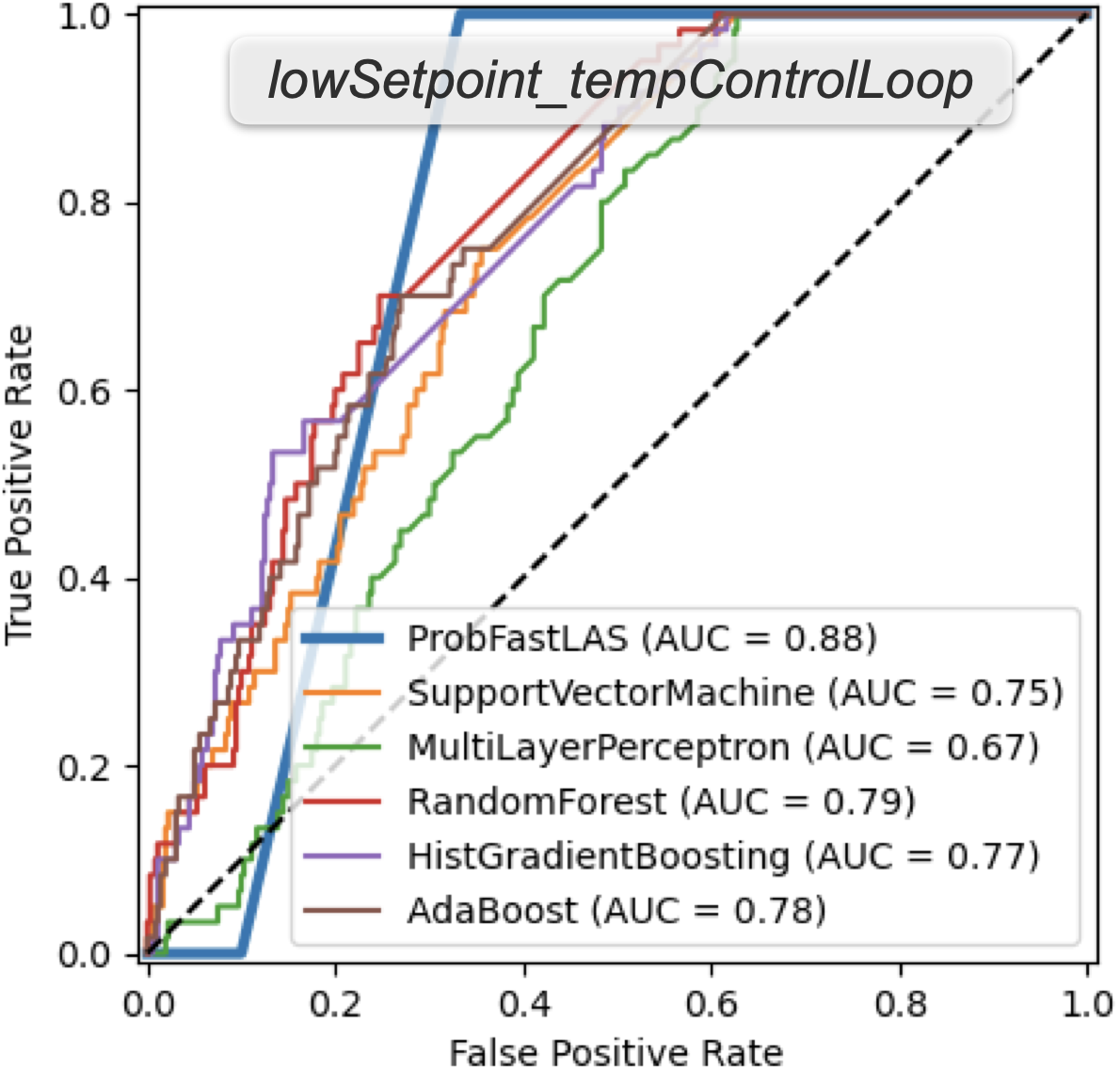}
  \par\vspace{2pt}
  \includegraphics[width=\linewidth]{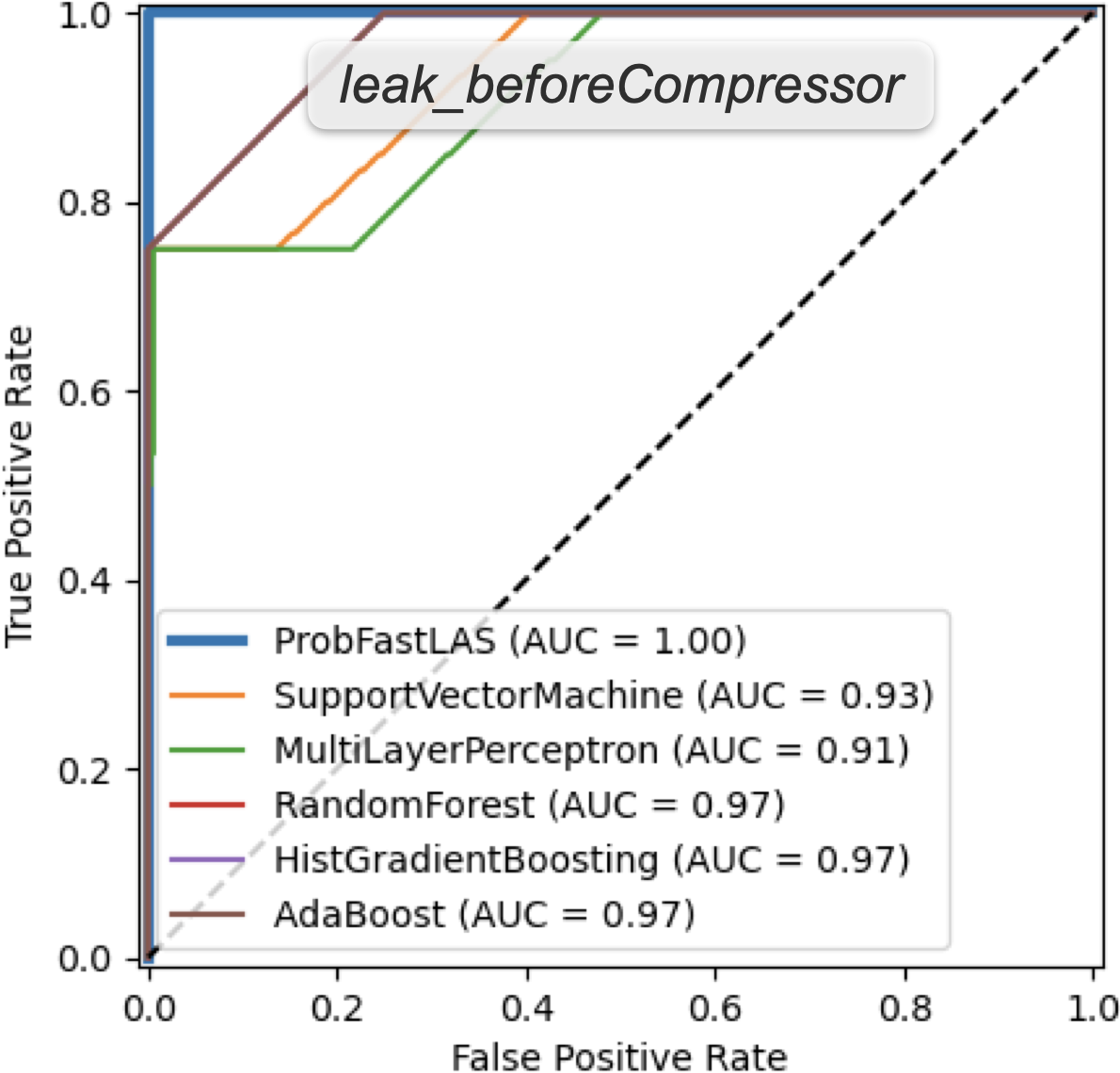}
\end{subfigure}
\begin{subfigure}[t]{0.3\linewidth}
  \centering
  \includegraphics[width=\linewidth]{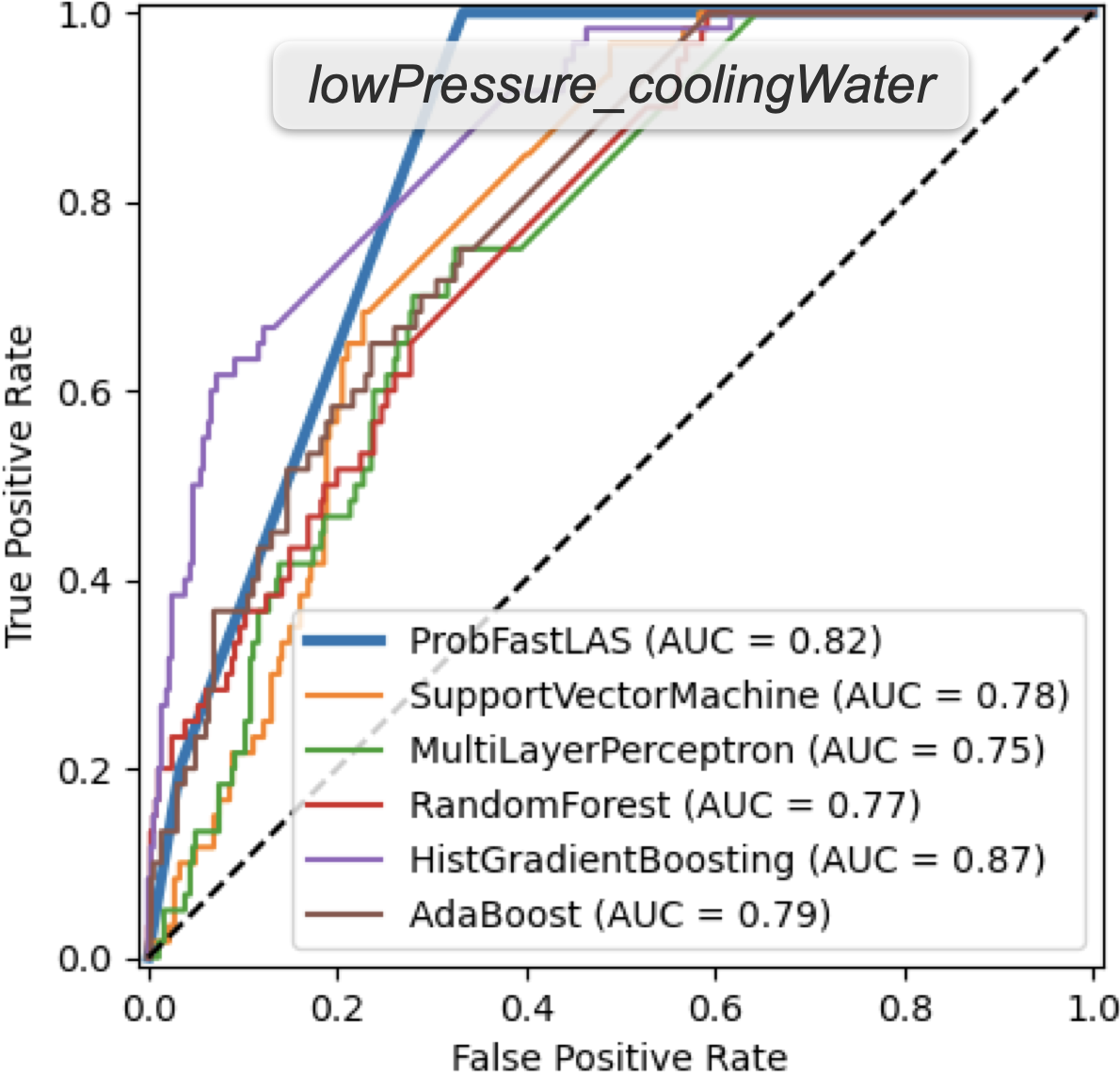}
  \par\vspace{2pt}
  \includegraphics[width=\linewidth]{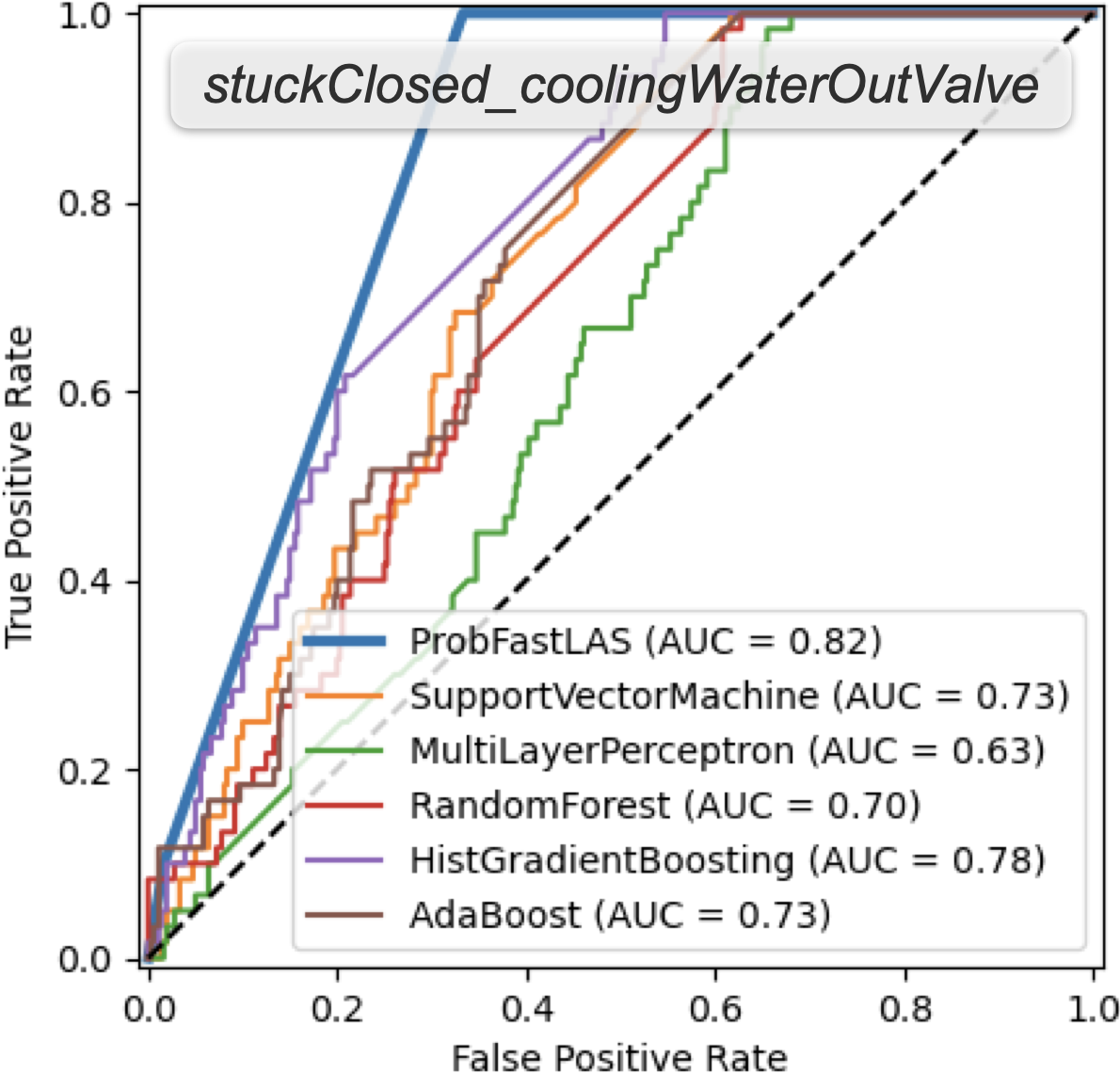}
\end{subfigure}
\caption{ROC curves generated from the default learning parameter configuration, for our approach and the baselines}
\label{fig:baseline_graphs}
\end{figure*}

Figure~\ref{fig:baseline_graphs} includes ROC curves from all five baselines, for each of the default failure cases. We can see that for both our approach and the baselines, there is are a large discrepancy in terms of AUC between failures, which is due to the fact that some failures result in similar process conditions, therefore making them difficult to distinguish.

Among the baselines, some approaches such as RF and AB, provide explainability through decision trees. In our case, the parent nodes of decision trees branch using numeric comparisons over process variable values (at $t_{short\_term}$, as with $T_{dynamic}$). This provides good explainability when only one decision tree is used and their depth is kept small, however doing so resulted in far worse classification performance. For instance, setting the maximum depth of decision trees to 4 for RF (down from 15) caused a reduction in the average AUC by around 0.1. For examples of generated decision trees, see the Supplementary Material.

\section{Discussion}

\subsection{Generated Rules}

In this subsection, we list rules generated by $T_{static}$ and $T_{dynamic}$ when using the default learning parameter settings, and provide an interpretation in each case. The following units and multipliers are assumed for numeric variables: (\texttt{srcr1\_p}, 10, bar), (\texttt{k1\_p1}, 100, bar), (\texttt{k1\_p2}, 1, bar), (\texttt{e2\_tti}, 1, ºC), (\texttt{m1\_pv}, 1, ºC).



The rules generated by $T_{static}$ can be divided into obvious rules, rules that can provide real insight into the process dynamics, and misleading rules. Rules from the first two categories are shown in Figure~\ref{fig:generated_rules_static} with comments to explain them. For example, on line 6, we can see that a leak before the compressor causes a high product (ethylene oxide) concentration; if this concentration reaches the lower flammability limit (2.6\% by volume for ethylene oxide~\citep*{kondo2011temperature,pubchem2025ethylene}), then the mixture of fluids becomes potentially flammable.

\begin{figure*}[!t]
\begin{cleanframebox}
  \Suppressnumber
    \begin{lstlisting}[frame=none]
% obvious rules|\Reactivatenumber{1}|
0.7:|{\ }|k1_p1(low2) :- srcr1_p(P), P |$\leq$| 9. |\Comment{\% low source pressure -> low pressure before k1}||\Suppressnumber|

% interesting rules|\Reactivatenumber{2}|
0.3:|{\ }|e2_tti(high1) :- k1_p1(P), P |$\geq$| 39, P |$\leq$| 46. |\Comment{\% low pressure -> more compression}|
1:|{\ }|r1_t2(low1) :- failure(beforeCompressor,leak). |\Comment{\% less flow -> less heat from reaction}|
1:|{\ }|r1_tau(high2) :- failure(beforeCompressor,leak). |\Comment{\% less flow -> longer residence time}|
1:|{\ }|r1_xmax(high1) :- failure(beforeCompressor,leak). |\Comment{\% ... -> higher conversion rate}|
1:|{\ }|snk1_z_c2h4o(high2) :- failure(beforeCompressor,leak). |\Comment{\% ... -> eo more concentrated}|
    \end{lstlisting}
\end{cleanframebox}
\caption{Rules generated by DisPLAS for $T_{static}$ ($\texttt{low2} < \texttt{low1} < \texttt{normal} < \texttt{high1} < \texttt{high2}$) when including all nontrivial experiments and process variable measurements over 125 runs of the simulation}
\label{fig:generated_rules_static}
\end{figure*}

\begin{figure*}[!t]
\begin{subfigure}[c]{\linewidth}
  \begin{cleanframebox}
        \begin{lstlisting}[frame=none]
1:|{\ }|failure(source,missingEthylene) :- m1_pv|$\nearrow$|(T), T |$\geq$| 3.
1:|{\ }|failure(beforeCompressor,leak) :- unchanged(m2_pv), e2_tti|$\nearrow$|(_). |\Comment{\% more compression}|
0.1:|{\ }|failure(coolingWaterOutValve,stuckClosed) :- unchanged(m1_pv), e2_tti(T), T |$\leq$| 295.
0.4:|{\ }|failure(...) :- unchanged(e2_tti), m1_pv(152). |\Comment{\% cooling water cannot escape}|
        \end{lstlisting}
  \end{cleanframebox}
  \caption{\( short\_term := 6 \)}
  \label{fig:generated_rules_dynamic_shortterm_6}
\end{subfigure}
\par\bigskip
\begin{subfigure}[c]{\linewidth}
  \Suppressnumber
  \begin{cleanframebox}
        \begin{lstlisting}[frame=none]
        |\vspace*{-\baselineskip}\Reactivatenumber{5}|
1:|{\ }|failure(source,missingEthylene) :- k1_p2|$\searrow$|(P), P |$\geq$| 3. |\Comment{\% new thermodynamic properties}|
1:|{\ }|failure(...) :- m1_pv(T), T |$\geq$| 154, T |$\leq$| 162. |\Comment{\% diverges more than (1)}|
1:|{\ }|failure(beforeCompressor,leak) :- unchanged(m2_pv), e2_tti|$\nearrow$|(_).
1:|{\ }|failure(coolingWaterOutValve,stuckClosed) :- m1_pv(T), T |$\geq$| 165, T |$\leq$| 166. |\Comment{\% like (4)}|
        \end{lstlisting}
  \end{cleanframebox}
  \caption{\( short\_term := 20 \)}
  \label{fig:generated_rules_dynamic_shortterm_20}
\end{subfigure}
\caption{Rules generated by DisPLAS for $T_{dynamic}$, modifying only the learning parameter $t_{short\_term}$ — the predicate \texttt{unchanged} indicates process variables with equal initial and short-term values, while remaining predicates in rule bodies refer to measurements taken at $t_{short\_term}$ and (absolute) changes in measurement since $t=0$ (increase: $\nearrow$, decrease: $\searrow$)}
\label{fig:generated_rules_dynamic}
\end{figure*}

In the dynamic case, the rules generated to identify three nontrivial failures are shown in Figure~\ref{fig:generated_rules_dynamic}, for two different values of $t_{short\_term}$. Comparing the sets of rules, we notice an increase in their confidence and usefulness with $t_{short\_term}$, as the repercussions of failures become progressively more pronounced. Note that these rules are rather general, owing to the limited pool of failures used for training. To generate rules that take into account the wide range of abnormal conditions possible in the real world, we would need to: model a much more complete selection of failures; or, fabricate training examples corresponding to ``unknown'' failures, by applying perturbations to existing examples and ensuring there is no overlap with any genuine examples.
Also, oscillation patterns are not currently being explicitly captured by the learning task; this is part of future work.

\subsection{Chemical Plant Agents}

Based on the learning tasks outlined in this paper — $T_{static}$ and $T_{dynamic}$ — we envisage a future agentic framework in which a ``family'' of collaborating agents is used to detect failures in a process. First, an agent $a_0$ is assigned the task $T_{static}$, and learns the cause-effect relations in the process from the static simulation. (Note that $a_0$ could also be used for validating a HAZOP study in the design phase of the process.) Alongside this agent, a set of agents $a_1,\dots,a_n$ learn failure detection rules using $T_{dynamic}$; these agents differ in their training data by the time windows being looked at, the portion of the process being focused on, and/or the included process variables. In addition, some agents may include real failure data in their training set when available. While the process is running, $a_1,\dots,a_n$ then get evaluated against live process data to generate failure predictions. These predictions are compared against each other: the greater the agreement between agents with respect to a given failure, the higher the confidence about this failure having occurred. When a failure is detected with sufficient confidence, it can be cross-checked with the rules of $a_0$. If this failure has occurred, the associated process data can be fed back to the agentic framework for training, thus improving the model.

The collaborative nature of the suggested approach increases the confidence with which failures are detected, helping with AI trustworthiness. At the same time, high explainability and interpretability are retained since we are putting together the results of agents that already have these properties. For this to work, the main framework proposed in this paper would have to be improved, namely in terms of scalability, and a mechanism for splitting a complex process into sub-processes would have to be devised. The suggested agentic approach can be classified as an AI-in-the-loop solution — where humans remain in control of a process but receive guidance from an AI — which is in line with Industry 5.0, an industrial paradigm that focuses on sustainability and calls for a collaborative relationship between humans and AI agents~\citep{arunthavanathan2024artificial}.

\section{Conclusion}

In this paper, we present a novel methodology involving a state-of-the-art probabilistic symbolic learner — DisPLAS — used to learn the cause-effect relations that govern specific chemical processes, based on sparse data from a process simulation. The proposed solution consists of two learning tasks: one focused on understanding the overall causality relationships in a process, and the other aimed at learning probabilistic failure detection rules. Our approach is shown to outperform neural network-based alternatives with regards to failure detection, while also being inherently explainable. We describe how these tasks could be used in the context of Industry 5.0, where human operators and agents work together towards maintaining safe operating conditions in an industrial plant.

So far, we have focused on a small case study — that of an ethylene oxidation process — with a limited number of failures, and have assumed that no two failures can occur simultaneously. However, processes in real-world plants are much more complex: in addition to optimizing the mode bias and structure of examples, future work should focus on radically improving scalability by decomposing the problem or by turning to a neuro-symbolic solution. This should allow tackling more complex problem settings, such as identifying simultaneous faults, which is significantly more challenging due to potential interdependency between faults.

\section{Acknowledgments}
This work is supported by the ProSafe doctoral network (call HORIZON-MSCA-2022-DN-01, project no. 101119358), with funding from UKRI (training grant reference no. EP/S023356/1).

\bibliography{aaai2026}

\end{document}